\pdfoutput=1
\documentclass[11pt]{article}

\usepackage{ACL2023}

\usepackage{times}
\usepackage{latexsym}

\usepackage[T1]{fontenc}
\usepackage[utf8]{inputenc}

\usepackage{microtype}

\usepackage{inconsolata}

\usepackage{amsmath,graphicx}
\usepackage{booktabs}

\title{Improving BERT with Hybrid Pooling Network and Drop Mask}

\author{Qian Chen, Wen Wang, Qinglin Zhang, Chong Deng, Ma Yukun, Siqi Zheng \\
  Speech Lab of DAMO Academy, Alibaba Group\\
\texttt{tanqing.cq,w.wang,qinglin.zql,dengchong.d,yukun.ma,zsq174630\}@alibaba-inc.com} }

\begin{document}
\maketitle
\begin{abstract}
Transformer-based pre-trained language models, such as BERT, achieve great success in various natural language understanding tasks. Prior research found that BERT captures a rich hierarchy of linguistic information at different layers. However, the vanilla BERT uses the same self-attention mechanism for each layer to model the different contextual features. In this paper, we propose a \emph{HybridBERT} model which combines self-attention and pooling networks to encode different contextual features in each layer. Additionally, we propose a simple \emph{DropMask} method to address the mismatch between pre-training and fine-tuning caused by excessive use of special mask tokens during Masked Language Modeling pre-training. Experiments show that HybridBERT outperforms BERT in pre-training with lower loss, faster training speed (8\% relative), lower memory cost (13\% relative), and also in transfer learning with 1.5\% relative higher accuracies on downstream tasks. Additionally, DropMask improves accuracies of BERT on downstream tasks across various masking rates.

\end{abstract}

\section{Introduction}
\label{sec:intro}
Transformer-based pre-trained language models, such as BERT~\cite{DBLP:conf/naacl/DevlinCLT19} and RoBERTa~\cite{DBLP:journals/corr/abs-1907-11692}, demonstrate remarkable success in various natural language understanding tasks. These models rely heavily on self-attention mechanisms to learn contextual features in natural language. Previous research~\cite{DBLP:conf/acl/JawaharSS19} shows that BERT captures a rich \emph{hierarchy} of linguistic information, encoding surface features at the bottom layers, syntactic features in the middle layers, and semantic features at the top layers. However, most works follow the BERT architecture and use the same self-attention mechanism to model different contextual features for each layer, neglecting the different contextual features from bottom to top layers. Moreover, it has been observed that a significant proportion of self-attention focuses on the previous/next token, the special mask token, or a combination of the two~\cite{DBLP:conf/emnlp/KovalevaRRR19}, indicating that self-attention for natural language learning may be redundant.

In this work, we aim to address the aforementioned hierarchy and redundancy issues in BERT, improve its transfer learning performance on downstream tasks, and reduce its training cost by accelerating training speed and lowering memory usage. Appendix~\ref{sec:related_work} summarizes related work. Prior attempts on stacking different model structures in different layers only achieve small gains, not yet reaching transferability of BERT. For example, mixing self-attention layer and Fourier Transform layer only achieves 97\%-99\% accuracy of BERT on downstream tasks~\cite{DBLP:conf/naacl/Lee-ThorpAEO22}. Prior works show that pooling networks can effectively model token mixing and long-range dependeices with \emph{linear time and memory complexity}. Particularly, the multi-granularity pooling network~\cite{DBLP:conf/iclr/TanCWZZL22} achieves 95.7\% accuracy of BERT on downstream tasks, showing a good tradeoff between time\slash memory efficiency and transfer learning ability. Hence, we propose a \textbf{HybridBERT} model to replace some self-attention layers in BERT with multi-granularity pooling network layers. We hypothesize that layer-wise mixing of self-attention (quadratic complexity) and pooling networks could simultaneously capture linguistic hierarchy and reduce redundancy.  
HybridBERT outperforms BERT by 1.5\% relative gains on average accuracy on downstream tasks. Also, since the multi-granularity pooling network has a linear complexity, HybridBERT has faster training speed and lower memory cost compared to BERT.

To further enhance transfer learning, we also aim to improve Masked Language Modeling (MLM)~\cite{DBLP:conf/naacl/DevlinCLT19}, which is crucial for the success of self-supervised learning of natural language. Typically, MLM replaces around 15\% of the input tokens with the special token [MASK]. This corruption creates a mismatch since the model sees the [MASK] token during pre-training but not during fine-tuning on downstream tasks.
ELECTRA~\cite{DBLP:conf/iclr/ClarkLLM20} addresses this mismatch by replacing some tokens with surrogates sampled from small generator networks; however, this approach requires additional generator networks\footnote{More related works are in Appendix~\ref{sec:related_work}.}. In this paper, we propose a simple and efficient \textbf{DropMask} method to address the mismatch in MLM.

Our contributions include:
(1) We propose a HybridBERT model that replaces some self-attention layers in BERT with pooling network layers.
(2) We propose a simple and efficient DropMask method to improve the transfer learning ability of BERT without adding parameters or compromising training speed and memory cost.
(3) Experiments show that HybridBERT outperforms BERT in pre-training with lower loss, faster training speed (8\% relative), lower memory cost (13\% rel.), and 5\% rel. higher average accuracy on downstream tasks. To the best of our knowledge, HybridBERT is the first model with different model structures in different layers that outperforms BERT in both reduced pre-training cost and improved transfer learning ability. DropMask improves accuracy of BERT on downstream tasks across various masking rates.

\section{Our Approach}

\label{sec:method}
\subsection{HybridBERT}
\label{sec:HybridBERT}
\begin{figure}[htb]
\centering
\includegraphics[width=4.5cm]{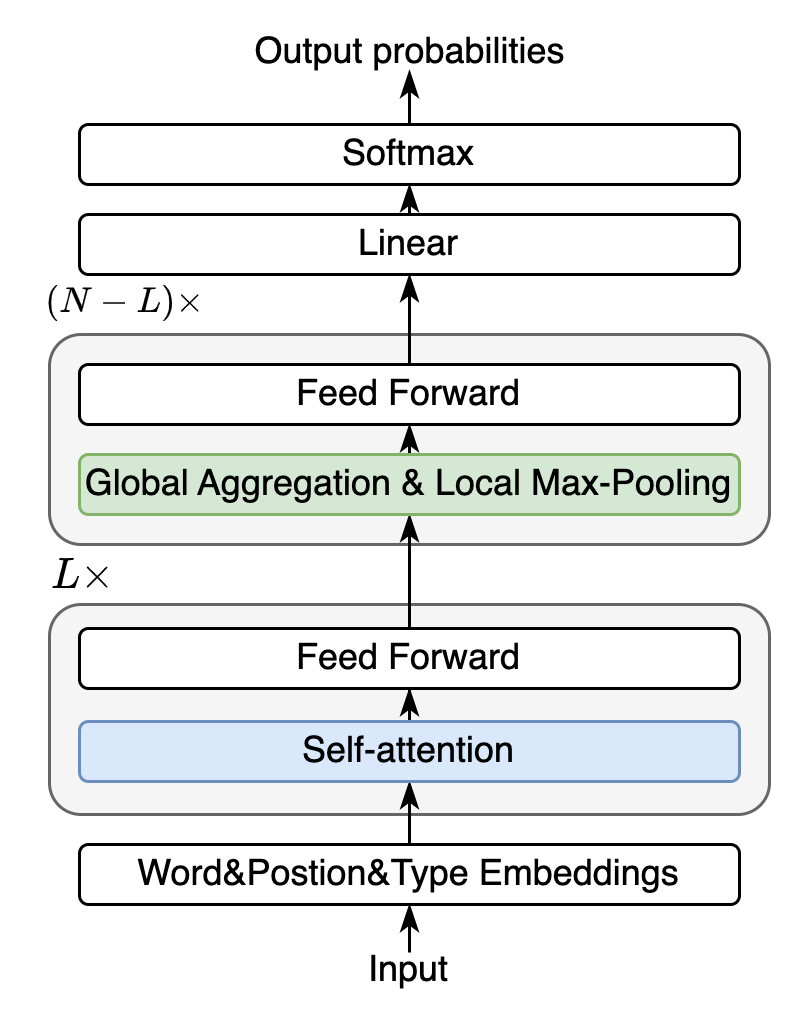}
\caption{The architecture of HybridBERT, where the self-attention layers are located at the bottom (near the input) and the pooling network layers are located at the top (near the output).}
\label{fig:hybrid_bert}
\end{figure}

Fig.~\ref{fig:hybrid_bert} illustrates the proposed HybridBERT. It consists of $L$ self-attention layers and $N-L$ pooling network layers (where $N$ is the total number of layers). First, the input sequence of tokens $\{x_1,\dots, x_l\}$ maps to word-, position-, and type-embeddings, where $l$ is the sequence length. These three different embeddings are then added as the input embeddings for the encoder, denoted by $H^{0} = \{h_1, \dots, h_l\}$. The self-attention layers are the same as in Transformer~\cite{DBLP:conf/nips/VaswaniSPUJGKP17}. 

\begin{figure}[htb]
\centering
\includegraphics[width=6cm]{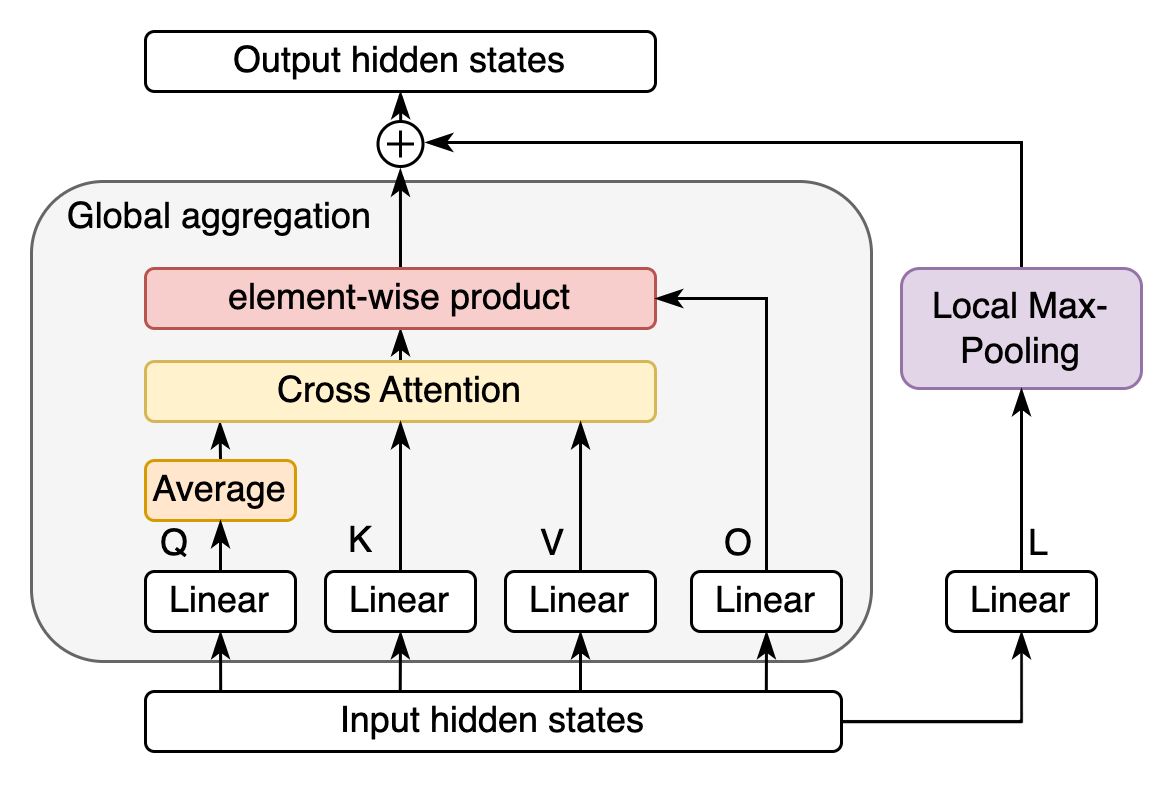}
\caption{The pooling network layers include global aggregation and local max-pooling.}
\label{fig:ga_lmp}
\end{figure}

To model structural dependencies at the top layers (near the output), we use a multi-granularity pooling network (PoNet)~\cite{DBLP:conf/iclr/TanCWZZL22}, which includes global aggregation (GA) and local max-pooling (LMP). We remove the original segment max-pooling in PoNet so that we align with BERT for generic scenarios, which does not rely on prerequisite structure in the source data.
Fig.~\ref{fig:ga_lmp} provides a detailed illustration of the pooling network layers. For the input hidden states to the pooling network layers, five different linear projections are applied: $ H_{*} = H_{in} W_{*}$, where $*$ can be $Q$, $K$, $V$, $O$, and $L$. Among them, $Q$, $K$, and $V$ denote query, key, and value; $O$ denotes the output fusion for global aggregation; $L$ denotes local max-pooling.

\paragraph{Global Aggregation}
Global Aggregation (GA) aims to capture the most important global information for each token and guarantee linear computational complexity. First, we average $H_Q$ to obtain a rough representation of the sequence $h^{avg}_Q$. Second, we take the rough representation $h^{avg}_Q$ as a query to perform multi-head cross-attention on the input sequence to obtain $h^{att}_Q$, as follows: 
$h^{att}_Q = \text{Attention}(h^{avg}_Q, H_K, H_V)$. 
Therefore, $h^{att}_Q$ provides a more accurate sequence representation compared to $h^{avg}_Q$. Note that $h^{avg}_Q$ is a single token, hence the computational complexity of this cross-attention is $O(N)$. 
To avoid all tokens sharing the same global representation of $h^{att}_Q$, the global aggregation output is the element-wise product between $h^{att}_Q$ and $H_O$: 
$H_{GA} = H_O \odot h^{att}_Q $. 

\paragraph{Local Max-Pooling}
Local Max-Pooling (LMP) is standard max-pooling over a sliding window to capture local contextual information for each token as: $H_{LMP} = \text{LMP}(H_L)$. The sliding window size and stride length are set to 3 and 1 in all our experiments. 
Finally, the output hidden states of the pooling network are the sum of these two features: $H_{out} = H_{GA}+H_{LMP}$.

\begin{table*}[htb]
\renewcommand{\arraystretch}{0.9}
\begin{center}
\scalebox{0.7}{
\begin{tabular}{l | c c c c c c c c}
\toprule
Model & AFQMC & CMNLI & CSL & IFLYTEK & QCNLI & TNEWS & WSC & \textbf{Avg} \\
\midrule
(1) BERT(12A) & 70.18\footnotesize{$\pm0.26$} & 74.39\footnotesize{$\pm0.26$} & \textbf{78.92}\footnotesize{$\pm0.73$}& 57.94\footnotesize{$\pm0.38$} & 68.69\footnotesize{$\pm0.30$} & 54.80\footnotesize{$\pm0.27$} & 66.45\footnotesize{$\pm0.70$} & 67.34\footnotesize{$\pm0.18$} \\
\underline{(2) HybridBERT(B8A+T4P)} & \textbf{70.95}\footnotesize{$\pm0.40$} & 75.30\footnotesize{$\pm0.23$} & 78.41\footnotesize{$\pm0.29$}& \textbf{58.82}\footnotesize{$\pm0.22$} & 69.70\footnotesize{$\pm0.14$} & 55.58\footnotesize{$\pm0.26$} & \textbf{69.80}\footnotesize{$\pm1.20$} & \textbf{68.37}\footnotesize{$\pm0.24$} \\
(3) HybridBERT(B4P+T8A) & 70.88\footnotesize{$\pm0.18$} & \textbf{75.45}\footnotesize{$\pm0.12$} & 78.57\footnotesize{$\pm0.27$}& 58.10\footnotesize{$\pm0.10$} & \textbf{69.71}\footnotesize{$\pm0.54$} & 55.39\footnotesize{$\pm0.20$} & 66.65\footnotesize{$\pm1.33$} & 67.82\footnotesize{$\pm0.20$} \\
(4) HybridBERT(B4A+T8P) & 70.34\footnotesize{$\pm0.13$} & 74.19\footnotesize{$\pm0.20$} & 78.64\footnotesize{$\pm0.45$}& 57.99\footnotesize{$\pm0.31$} & 68.73\footnotesize{$\pm0.11$} & 55.63\footnotesize{$\pm0.21$} & 66.84\footnotesize{$\pm1.65$} & 67.48\footnotesize{$\pm0.22$} \\ 
(5) HybridBERT(B8P+T4A) & 70.57\footnotesize{$\pm0.30$} & 75.40\footnotesize{$\pm0.32$} & 77.79\footnotesize{$\pm0.38$}& 57.92\footnotesize{$\pm0.23$} & 69.16\footnotesize{$\pm0.29$} & 55.32\footnotesize{$\pm0.22$} & 66.71\footnotesize{$\pm1.44$} & 67.55\footnotesize{$\pm0.36$} \\ 
(6) PoNet(12P) & 68.99\footnotesize{$\pm0.17$} & 72.50\footnotesize{$\pm0.19$} & 74.07\footnotesize{$\pm0.55$}& 57.90\footnotesize{$\pm0.23$} & 67.07\footnotesize{$\pm0.53$} & \textbf{55.69}\footnotesize{$\pm0.09$} & 64.41\footnotesize{$\pm0.94$} & 65.80\footnotesize{$\pm0.16$} \\
\midrule
(7) HybridBERT(B8A+T4P) w/o GA  & 70.61\footnotesize{$\pm0.38$} & 74.63\footnotesize{$\pm0.40$} & 78.63\footnotesize{$\pm0.37$}& 58.01\footnotesize{$\pm0.29$} & 69.07\footnotesize{$\pm0.30$} & 55.02\footnotesize{$\pm0.25$} & 70.92\footnotesize{$\pm0.97$} & 68.13\footnotesize{$\pm0.16$} \\ 
(8) HybridBERT(B8A+T4P) w/o LMP  & 69.80\footnotesize{$\pm0.17$} & 75.76\footnotesize{$\pm0.09$} & 78.73\footnotesize{$\pm0.44$}& 58.06\footnotesize{$\pm0.18$} & 69.92\footnotesize{$\pm0.23$} & 55.52\footnotesize{$\pm0.23$} & 67.63\footnotesize{$\pm0.89$} & 67.90\footnotesize{$\pm0.19$} \\ 
\midrule
(9) BERT(12A) w/ DM & 70.67\footnotesize{$\pm0.50$} & 75.34\footnotesize{$\pm0.13$} & 78.83\footnotesize{$\pm0.21$}& 58.49\footnotesize{$\pm0.28$} & 68.88\footnotesize{$\pm0.30$} & 55.07\footnotesize{$\pm0.09$} & 67.11\footnotesize{$\pm0.73$} & 67.77\footnotesize{$\pm0.21$} \\ 
\bottomrule
\end{tabular}
}
\end{center}
\caption{Comparison of various models on the Chinese CLUE development set. The metric for each dataset is Accuracy. \textbf{Avg} is the average accuracy across all datasets. The means and standard deviations are reported from 5 runs with different random seeds. We conduct grid search on hyperparameters for each seed to obtain its best results. }
\label{tab:clue}
\end{table*}

\subsection{DropMask}
\label{sec:drop_mask}

Masked Language Modeling is the crucial training objective for the success of BERT and BERT-like models.
Typically, the masking rate is set as 15\%. 80\% of the masked positions are replaced with the special token [MASK], 10\% are replaced with randomly sampled words, and the remaining 10\% are left unchanged. 
Having too many [MASK] tokens in input sequence can cause significant mismatch between pre-training and fine-tuning, as downstream tasks do not have [MASK] in their input sequence.
The recently proposed MAE architecture~\cite{DBLP:conf/cvpr/HeCXLDG22} uses an asymmetric encoder-decoder to process a subset of visible image patches, thereby enabling training large encoders with small computation cost. Inspired by this approach, we propose a simple and efficent DropMask method for self-attention in BERT models. In contrast to MAE, which removes masked tokens from the encoder, DropMask only excludes [MASK] tokens from calculation of weighted summation for self-attention, as follows:
\begin{equation} 
 \label{eq:out}
 y_j = \sum_{i \neq \text{[MASK]}}{A_{ij} x_{i}},~~~i,j \in \{1,\dots, l\}
 \end{equation}
where $A$ is the dot-product attention matrix, $x$ is the input value, and $y$ is the output of weighted summation. 
DropMask enables [MASK] tokens to view other unmasked tokens, while preventing all tokens from viewing [MASK] tokens, thereby reducing mismatch between pre-training and fine-tuning in BERT.

\section{Experiments}
\label{sec:experments}
% \subsection{Implementation}
We conduct pre-training of various models and evaluate their fine-tuning performance on the English GLUE benchmark~\cite{DBLP:conf/iclr/WangSMHLB19} and the Chinese CLUE benchmark~\cite{DBLP:conf/coling/XuHZLCLXSYYTDLS20}. The implementation details are in Appendix~\ref{sec:implementation}. 

\begin{table*}[htb]
\renewcommand{\arraystretch}{0.8}
\begin{center}
\scalebox{0.7}{
\begin{tabular}{l | c  | c c c c c c c c c}
\toprule
Model & Masking Rate  & MNLI & QNLI & QQP & SST-2 & CoLA & MRPC & RTE & STS-B & \textbf{Avg} \\
\midrule
BERT & 5\%  & 82.99 & 90.21 & 89.35 & 91.97 & 55.27 & 86.69 & 57.76 & 81.17 & 79.93 \\
BERT w/ DM & 5\%    & 82.29 & 89.44 & 89.22 & 92.09 & 56.77 & 88.91 & 59.95 & 86.95 & 80.70 \\
\midrule
BERT & 15\%    & 83.45 & 90.24 & 89.60 & \textbf{92.55} & 60.61 & 87.75 & 64.26 & 87.81 & 82.03 \\
BERT w/ DM&  15\%   & 82.96 & 90.30 & 89.55 & 92.43 & \textbf{60.82} & 88.66 & \textbf{65.70} & \textbf{88.44} & \textbf{82.36} \\
\midrule
BERT &  30\%   & \textbf{83.62} & 89.51 & 89.49 & 91.86 & 57.79 & 89.22 & 62.45 & 88.11 & 81.51 \\
BERT w/ DM &  30\%   & 83.54 & \textbf{90.48} & \textbf{89.61} & 92.43 & 60.34 & 88.50 & 64.26 & 88.05 & 82.15 \\
\midrule
BERT & 75\%    & 81.43 & 88.36 & 89.21 & 90.48 & 32.13 & 86.35 & 58.48 & 84.89 & 76.41 \\
BERT w/ DM & 75\% & 81.18 & 88.39 & 89.15 & 91.40 & 50.52 & 86.40 & 58.84 & 86.70 & 79.08 \\
\bottomrule
\end{tabular}
}
\end{center}
\caption{Performance of various models on the English GLUE validation set. The best GLUE results for each model are reported using grid search on hyperparameters. The reported metrics include: the mean accuracy on match/mismatch splits for MNLI, accuracy for QNLI, SST-2, RTE, Matthews correlation coefficient for CoLA, the mean of Pearson correlation coefficient and Spearman's rank correlation coefficient for STS-B, and the mean of accuracy and F1 for QQP and MRPC.
}
\label{tab:glue}
\end{table*}

\begin{table}[htb]
\renewcommand{\arraystretch}{0.8}
\begin{center}
\scalebox{0.7}{
\begin{tabular}{c | c c c | c c  }
\toprule
Model& $L_{total}$$\downarrow$ & $L_{MLM}$$\downarrow$ & $L_{SSO}$$\downarrow$ & Speed$\downarrow$ & Memory$\downarrow$\\
\midrule
(1) &1.556 & 1.344 & 0.212 &  1.44 & 81.0 \\
(2) & \textbf{1.525} & \textbf{1.301} & 0.224 &  1.32 & 70.5 \\
(3) & 1.542 & 1.335 & \textbf{0.207} & 1.23 & 59.8 \\
(4) &1.608 & 1.371 & 0.237 & 1.23 & 59.8 \\
(5) & 1.601 & 1.388 & 0.213 &  1.23 & 60.0 \\
(6) & 1.911 & 1.651 & 0.260 &  \textbf{1.12} & \textbf{49.2} \\
\midrule
(7) & 1.499 & 1.278 & 0.221 & 1.27 & 68.8 \\
(8) & 1.498 & 1.288 & 0.210 & 1.30 & 68.7 \\
\midrule
(9) & 1.563 & 1.345 & 0.219 & 1.44 & 81.3 \\
\bottomrule
\end{tabular}
}
\end{center}
\caption{Comparison of validation set evaluation losses during pre-training and training speed and memory cost on Chinese Wikipedia for all the indexed models in Table~\ref{tab:clue}, using 4 Tesla V100 cards. The training speed is measured in seconds per iteration, and the memory cost is measured in total GB of the 4 Tesla V100 cards.}
\label{tab:loss}
\end{table}

\noindent \textbf{Results of HybridBERT on CLUE}
We evaluate different configurations of HybridBERT and applying DropMask to BERT on the CLUE development set (see able \ref{tab:clue}). We focus on the average score (\textbf{Avg}) across 7 datasets.
The first baseline uses 12 self-attention layers (denoted as BERT(12A)) and the second baseline uses 12 pooling network layers (denoted as PoNet(12P)). We evaluate the impact from varying the number of self-attention layers and the placement of self-attention layers and pooling network layers in HybridBERT.  We make two observations: (1) 8 self-attention layers perform better than 4 self-attention layers; (2) placing the self-attention layer at the bottom  (near the input) yields better or comparable results, which is opposite to observations on FNet-Hybrid~\cite{DBLP:conf/naacl/Lee-ThorpAEO22}. FNet-Hybrid finds that placing the self-attention layers at the top (near the output) achieves better accuracy on downstream tasks than at the bottom.
The different observations may be due to the distinct mechanisms used by Fourier Transform layer and pooling network.
The best-performing configuration for HybridBERT places 8 self-attention layers at the bottom and 4 pooling layers at the top, denoted as \textbf{Hybrid(B8A+T4P)}. It improves BERT by \textbf{1.03} absolute and PoNet by \textbf{2.57} absolute on the Avg score. Notably, Hybrid(B8A+T4P) outperforms BERT on all CLUE tasks except CSL.  We focus on the best-performing HybridBERT(B8A+T4P) for ablation analysis. The Avg score drops -0.24 from removing GA from this model and -0.47 from removing LMP, suggesting that both GA and LMP are important for good performance on downstream tasks. 

\noindent \textbf{Results of HybridBERT on Pre-training}
We compare the performance of different models on total loss ($L_{total}$), MLM loss ($L_{MLM}$), and Sentence Structural Objective (SSO)~\cite{DBLP:conf/iclr/TanCWZZL22} loss ($L_{SSO}$) on validation set, as well as the training speed and memory cost, during pre-training (see Table \ref{tab:loss}).
Compared with BERT, our HybridBERT(B8A+T4P) improves $L_{total}$ by 0.031 and $L_{MLM}$ by 0.043 but degrades $L_{SSO}$ by 0.012. Compared with PoNet, HybridBERT(B8A+T4P) improves $L_{total}$ by 0.386, $L_{MLM}$ by 0.35, and $L_{SSO}$ by 0.036. 

Meanwhile, HybridBERT(B8A+T4P) has faster training speed (8\% relative) and less memory cost (13\% rel.) compared to BERT, but slower training speed (18\% rel.) and more memory cost (43\% rel.) than PoNet.
The effect of changing the number of self-attention layers and self-attention positions in HybridBERT on the evaluation loss during pre-training is consistent with the performance on CLUE. More self-attention layers increase accuracy at the cost of training speed and memory cost. Self-attention positions have no significant impact in training speed and memory cost.
For ablation analysis on removing GA and LMP, both HybridBERT w/o GA and HybridBERT w/o LMP outperform HybridBERT on $L_{total}$ and have faster training speed and less memory cost. However, they do not perform as well as HybridBERT on downstream CLUE datasets (see Table \ref{tab:clue}). 

\noindent \textbf{Results of DropMask}
We evaluate BERT w/ or w/o DropMask (DM) on CLUE (Table \ref{tab:clue}) and GLUE (Table \ref{tab:glue}).
BERT w/DM achieves \textbf{0.44} absolute gain on CLUE development set and \textbf{0.33} abs. gain on GLUE validation set.
Table \ref{tab:glue} shows that BERT w/DM achieves \textbf{[0.33, 2.67]} abs. gains on GLUE validation set across different masking rates.
% , except for a 50\% masking rate. 
Interestingly, with 75\% masking rate, w/DM achieves a large \textbf{18.39} abs. gain on the COLA dataset.
We also add DropMask to HybridBERT but observe no improvement. We plan to investigate this in future work. Table~\ref{tab:loss} verifies that DM does not change training speed and memory cost\footnote{The very small difference between Model (1) and (9) is due to different implementations.}. 

\section{Conclusion}
\label{sec:cnclusion}
We propose a HybridBERT model and a DropMask method to improve BERT accuracy on downstream tasks. HybridBERT outperforms BERT in pre-training cost and in higher accuracy on downstream tasks. DropMask improves the accuracy of BERT on downstream tasks across different masking rates.
Future work includes investigating other efficient Transformers in hybrid structures. 
\newpage

\section*{Limitations}
The study has several limitations to consider. Firstly, our experiments are conducted only on the BERT-base size model, so it is uncertain whether the results would hold for larger models. Secondly, we evaluate the Hybrid pooling networks and DropMask only on the BERT backbone, not yet including other backbones such as RoBERTa. Lastly, we have not observed improvements from applying DropMask on HybridBERT, which requires further investigations.

\bibliography{custom}

\begin{thebibliography}{21}
\expandafter\ifx\csname natexlab\endcsname\relax\def\natexlab#1{#1}\fi

\bibitem[{Clark et~al.(2020)Clark, Luong, Le, and
  Manning}]{DBLP:conf/iclr/ClarkLLM20}
Kevin Clark, Minh{-}Thang Luong, Quoc~V. Le, and Christopher~D. Manning. 2020.
\newblock \href {https://openreview.net/forum?id=r1xMH1BtvB} {{ELECTRA:}
  pre-training text encoders as discriminators rather than generators}.
\newblock In \emph{{ICLR},}. OpenReview.net.

\bibitem[{Devlin et~al.(2019)Devlin, Chang, Lee, and
  Toutanova}]{DBLP:conf/naacl/DevlinCLT19}
Jacob Devlin, Ming{-}Wei Chang, Kenton Lee, and Kristina Toutanova. 2019.
\newblock \href {https://doi.org/10.18653/v1/n19-1423} {{BERT:} pre-training of
  deep bidirectional transformers for language understanding}.
\newblock In \emph{{NAACL-HLT}}, pages 4171--4186. Association for
  Computational Linguistics.

\bibitem[{Gulati et~al.(2020)Gulati, Qin, Chiu, Parmar, Zhang, Yu, Han, Wang,
  Zhang, Wu, and Pang}]{DBLP:conf/interspeech/GulatiQCPZYHWZW20}
Anmol Gulati, James Qin, Chung{-}Cheng Chiu, Niki Parmar, Yu~Zhang, Jiahui Yu,
  Wei Han, Shibo Wang, Zhengdong Zhang, Yonghui Wu, and Ruoming Pang. 2020.
\newblock \href {https://doi.org/10.21437/Interspeech.2020-3015} {Conformer:
  Convolution-augmented transformer for speech recognition}.
\newblock In \emph{Interspeech}, pages 5036--5040. {ISCA}.

\bibitem[{He et~al.(2022{\natexlab{a}})He, Chen, Xie, Li, Doll{\'{a}}r, and
  Girshick}]{DBLP:conf/cvpr/HeCXLDG22}
Kaiming He, Xinlei Chen, Saining Xie, Yanghao Li, Piotr Doll{\'{a}}r, and
  Ross~B. Girshick. 2022{\natexlab{a}}.
\newblock \href {https://doi.org/10.1109/CVPR52688.2022.01553} {Masked
  autoencoders are scalable vision learners}.
\newblock In \emph{{CVPR}}, pages 15979--15988. {IEEE}.

\bibitem[{He et~al.(2022{\natexlab{b}})He, Peng, Lu, Wang, Mei, Liu, Xu,
  Awadalla, Shi, Zhu, Xiong, Zeng, Gao, and
  Huang}]{DBLP:journals/corr/abs-2208-09770}
Pengcheng He, Baolin Peng, Liyang Lu, Song Wang, Jie Mei, Yang Liu, Ruochen Xu,
  Hany~Hassan Awadalla, Yu~Shi, Chenguang Zhu, Wayne Xiong, Michael Zeng,
  Jianfeng Gao, and Xuedong Huang. 2022{\natexlab{b}}.
\newblock \href {https://doi.org/10.48550/arXiv.2208.09770} {Z-code++: {A}
  pre-trained language model optimized for abstractive summarization}.
\newblock \emph{CoRR}, abs/2208.09770.

\bibitem[{Jawahar et~al.(2019)Jawahar, Sagot, and
  Seddah}]{DBLP:conf/acl/JawaharSS19}
Ganesh Jawahar, Beno{\^{\i}}t Sagot, and Djam{\'{e}} Seddah. 2019.
\newblock \href {https://doi.org/10.18653/v1/p19-1356} {What does {BERT} learn
  about the structure of language?}
\newblock In \emph{{ACL}}, pages 3651--3657. Association for Computational
  Linguistics.

\bibitem[{Jiang et~al.(2020)Jiang, Yu, Zhou, Chen, Feng, and
  Yan}]{DBLP:conf/nips/JiangYZCFY20}
Zihang Jiang, Weihao Yu, Daquan Zhou, Yunpeng Chen, Jiashi Feng, and Shuicheng
  Yan. 2020.
\newblock \href
  {https://proceedings.neurips.cc/paper/2020/hash/96da2f590cd7246bbde0051047b0d6f7-Abstract.html}
  {Convbert: Improving {BERT} with span-based dynamic convolution}.
\newblock In \emph{NeurIPS}.

\bibitem[{Kovaleva et~al.(2019)Kovaleva, Romanov, Rogers, and
  Rumshisky}]{DBLP:conf/emnlp/KovalevaRRR19}
Olga Kovaleva, Alexey Romanov, Anna Rogers, and Anna Rumshisky. 2019.
\newblock \href {https://doi.org/10.18653/v1/D19-1445} {Revealing the dark
  secrets of {BERT}}.
\newblock In \emph{{EMNLP-IJCNLP}}, pages 4364--4373. Association for
  Computational Linguistics.

\bibitem[{Lee{-}Thorp et~al.(2022)Lee{-}Thorp, Ainslie, Eckstein, and
  Onta{\~{n}}{\'{o}}n}]{DBLP:conf/naacl/Lee-ThorpAEO22}
James Lee{-}Thorp, Joshua Ainslie, Ilya Eckstein, and Santiago
  Onta{\~{n}}{\'{o}}n. 2022.
\newblock \href {https://doi.org/10.18653/v1/2022.naacl-main.319} {Fnet: Mixing
  tokens with fourier transforms}.
\newblock In \emph{{NAACL}}, pages 4296--4313. Association for Computational
  Linguistics.

\bibitem[{Liu et~al.(2019)Liu, Ott, Goyal, Du, Joshi, Chen, Levy, Lewis,
  Zettlemoyer, and Stoyanov}]{DBLP:journals/corr/abs-1907-11692}
Yinhan Liu, Myle Ott, Naman Goyal, Jingfei Du, Mandar Joshi, Danqi Chen, Omer
  Levy, Mike Lewis, Luke Zettlemoyer, and Veselin Stoyanov. 2019.
\newblock \href {http://arxiv.org/abs/1907.11692} {Roberta: {A} robustly
  optimized {BERT} pretraining approach}.
\newblock \emph{CoRR}, abs/1907.11692.

\bibitem[{Peng et~al.(2022)Peng, Dalmia, Lane, and
  Watanabe}]{DBLP:conf/icml/PengDL022}
Yifan Peng, Siddharth Dalmia, Ian~R. Lane, and Shinji Watanabe. 2022.
\newblock \href {https://proceedings.mlr.press/v162/peng22a.html}
  {Branchformer: Parallel mlp-attention architectures to capture local and
  global context for speech recognition and understanding}.
\newblock In \emph{{ICML}}, volume 162 of \emph{Proceedings of Machine Learning
  Research}, pages 17627--17643. {PMLR}.

\bibitem[{Song et~al.(2020)Song, Tan, Qin, Lu, and
  Liu}]{DBLP:conf/nips/Song0QLL20}
Kaitao Song, Xu~Tan, Tao Qin, Jianfeng Lu, and Tie{-}Yan Liu. 2020.
\newblock \href
  {https://proceedings.neurips.cc/paper/2020/hash/c3a690be93aa602ee2dc0ccab5b7b67e-Abstract.html}
  {Mpnet: Masked and permuted pre-training for language understanding}.
\newblock In \emph{NeurIPS}.

\bibitem[{Tan et~al.(2022)Tan, Chen, Wang, Zhang, Zheng, and
  Ling}]{DBLP:conf/iclr/TanCWZZL22}
Chao{-}Hong Tan, Qian Chen, Wen Wang, Qinglin Zhang, Siqi Zheng, and Zhen{-}Hua
  Ling. 2022.
\newblock \href {https://openreview.net/forum?id=9jInD9JjicF} {Ponet: Pooling
  network for efficient token mixing in long sequences}.
\newblock In \emph{{ICLR}}. OpenReview.net.

\bibitem[{Vaswani et~al.(2017)Vaswani, Shazeer, Parmar, Uszkoreit, Jones,
  Gomez, Kaiser, and Polosukhin}]{DBLP:conf/nips/VaswaniSPUJGKP17}
Ashish Vaswani, Noam Shazeer, Niki Parmar, Jakob Uszkoreit, Llion Jones,
  Aidan~N. Gomez, Lukasz Kaiser, and Illia Polosukhin. 2017.
\newblock \href
  {https://proceedings.neurips.cc/paper/2017/hash/3f5ee243547dee91fbd053c1c4a845aa-Abstract.html}
  {Attention is all you need}.
\newblock In \emph{{NeurIPS}}, pages 5998--6008.

\bibitem[{Wang et~al.(2019)Wang, Singh, Michael, Hill, Levy, and
  Bowman}]{DBLP:conf/iclr/WangSMHLB19}
Alex Wang, Amanpreet Singh, Julian Michael, Felix Hill, Omer Levy, and
  Samuel~R. Bowman. 2019.
\newblock \href {https://openreview.net/forum?id=rJ4km2R5t7} {{GLUE:} {A}
  multi-task benchmark and analysis platform for natural language
  understanding}.
\newblock In \emph{{ICLR}}. OpenReview.net.

\bibitem[{Wang et~al.(2020)Wang, Bi, Yan, Wu, Xia, Bao, Peng, and
  Si}]{DBLP:conf/iclr/0225BYWXBPS20}
Wei Wang, Bin Bi, Ming Yan, Chen Wu, Jiangnan Xia, Zuyi Bao, Liwei Peng, and
  Luo Si. 2020.
\newblock \href {https://openreview.net/forum?id=BJgQ4lSFPH} {Structbert:
  Incorporating language structures into pre-training for deep language
  understanding}.
\newblock In \emph{{ICLR},}. OpenReview.net.

\bibitem[{Wettig et~al.(2022)Wettig, Gao, Zhong, and
  Chen}]{DBLP:journals/corr/abs-2202-08005}
Alexander Wettig, Tianyu Gao, Zexuan Zhong, and Danqi Chen. 2022.
\newblock \href {http://arxiv.org/abs/2202.08005} {Should you mask 15{\%} in
  masked language modeling?}
\newblock \emph{CoRR}, abs/2202.08005.

\bibitem[{Wu et~al.(2020)Wu, Liu, Lin, Lin, and Han}]{DBLP:conf/iclr/WuLLLH20}
Zhanghao Wu, Zhijian Liu, Ji~Lin, Yujun Lin, and Song Han. 2020.
\newblock \href {https://openreview.net/forum?id=ByeMPlHKPH} {Lite transformer
  with long-short range attention}.
\newblock In \emph{{ICLR}}. OpenReview.net.

\bibitem[{Xu et~al.(2020)Xu, Hu, Zhang, Li, Cao, Li, Xu, Sun, Yu, Yu, Tian,
  Dong, Liu, Shi, Cui, Li, Zeng, Wang, Xie, Li, Patterson, Tian, Zhang, Zhou,
  Liu, Zhao, Zhao, Yue, Zhang, Yang, Richardson, and
  Lan}]{DBLP:conf/coling/XuHZLCLXSYYTDLS20}
Liang Xu, Hai Hu, Xuanwei Zhang, Lu~Li, Chenjie Cao, Yudong Li, Yechen Xu, Kai
  Sun, Dian Yu, Cong Yu, Yin Tian, Qianqian Dong, Weitang Liu, Bo~Shi, Yiming
  Cui, Junyi Li, Jun Zeng, Rongzhao Wang, Weijian Xie, Yanting Li, Yina
  Patterson, Zuoyu Tian, Yiwen Zhang, He~Zhou, Shaoweihua Liu, Zhe Zhao, Qipeng
  Zhao, Cong Yue, Xinrui Zhang, Zhengliang Yang, Kyle Richardson, and Zhenzhong
  Lan. 2020.
\newblock \href {https://doi.org/10.18653/v1/2020.coling-main.419} {{CLUE:} {A}
  chinese language understanding evaluation benchmark}.
\newblock In \emph{{COLING}}, pages 4762--4772. International Committee on
  Computational Linguistics.

\bibitem[{Yang et~al.(2019)Yang, Dai, Yang, Carbonell, Salakhutdinov, and
  Le}]{DBLP:conf/nips/YangDYCSL19}
Zhilin Yang, Zihang Dai, Yiming Yang, Jaime~G. Carbonell, Ruslan Salakhutdinov,
  and Quoc~V. Le. 2019.
\newblock \href
  {https://proceedings.neurips.cc/paper/2019/hash/dc6a7e655d7e5840e66733e9ee67cc69-Abstract.html}
  {Xlnet: Generalized autoregressive pretraining for language understanding}.
\newblock In \emph{NeurIPS}, pages 5754--5764.

\bibitem[{Yu et~al.(2022)Yu, Luo, Zhou, Si, Zhou, Wang, Feng, and
  Yan}]{DBLP:conf/cvpr/YuLZSZWFY22}
Weihao Yu, Mi~Luo, Pan Zhou, Chenyang Si, Yichen Zhou, Xinchao Wang, Jiashi
  Feng, and Shuicheng Yan. 2022.
\newblock \href {https://doi.org/10.1109/CVPR52688.2022.01055} {Metaformer is
  actually what you need for vision}.
\newblock In \emph{{CVPR}}, pages 10809--10819. {IEEE}.

\end{thebibliography}
\bibliographystyle{acl_natbib}

\appendix

\section{Related work}
\label{sec:related_work}
\subsection{Transformers Variants}
Many previous works have studied improving Transformers by combining self-attention with convolutional networks to enhance context-dependent modeling at each layer. Conformer~\cite{DBLP:conf/interspeech/GulatiQCPZYHWZW20} inserts a convolutional sublayer between the self-attention sublayer and the feed-forward sublayer, which is found effective in many speech-processing tasks. ConvBERT~\cite{DBLP:conf/nips/JiangYZCFY20} replaces some self-attention heads with span-based dynamic convolutions to directly model local dependencies for improving performance on natural language understanding tasks. Similar two-branch architectures are used in Lite-Transformer~\cite{DBLP:conf/iclr/WuLLLH20} and Branchformer~\cite{DBLP:conf/icml/PengDL022}.

Meanwhile, some other works choose to stack different model structures in different layers. FNet-hybrid~\cite{DBLP:conf/naacl/Lee-ThorpAEO22} found that replacing the self-attention in the 10 bottom layers (near the input) with a Fourier transform layer runs 40\%-70\% faster, but achieves only up to 97\%-99\% of the accuracy of BERT on the GLUE benchmark~\cite{DBLP:conf/iclr/WangSMHLB19}. Furthermore, they found that placing the self-attention layers at the top (near the output) gave better accuracy on downstream tasks than the bottom.
Fusion-in-encoder (FiE)~\cite{DBLP:journals/corr/abs-2208-09770} uses a local layer at the bottom and a global layer (i.e. original self-attention) at the top to encode long sequences while preserving high attention precision on short sequences. In each local layer, the input sequence is divided into small patches, and self-attention is applied only to those patches locally.

Additionally, pooling mechanisms have recently shown great potential for token mixing and modeling long-range dependencies. PoNet~\cite{DBLP:conf/iclr/TanCWZZL22} proposes a pooling network for token mixing in long sequences with linear complexity. They design a multi-granularity pooling module to capture local, segmental and global contextual information and interactions. However, they only observe 95.7\% accuracy of BERT on the GLUE benchmark. PoolFormer~\cite{DBLP:conf/cvpr/YuLZSZWFY22} achieves competitive performance on multiple computer vision tasks using a simple spatial pooling operator.
In this paper, considering the rich hierarchy of linguistic information, we propose a HybridBERT model that replaces some of the self-attention layers in BERT with pooling networks.

\subsection{Self-supervised Pre-training Tasks}
Masked language modeling (MLM) is one of the most successful self-supervised pre-training tasks for natural language processing, which is used in BERT, RoBERTa, and other BERT-like models.
% BERT uses masked language modeling (MLM) for pre-training and is one of the most successful pre-training models. 
It corrupts the input token sequence by randomly replacing some tokens with [MASK] and then trains a model to reconstruct the original tokens. However, this corruption process leads to mismatches in BERT, where the model sees the [MASK] token during pre-training, but not when fine-tuning on downstream tasks.
There has been much effort to address the issues of MLM. XLNet~\cite{DBLP:conf/nips/YangDYCSL19} introduces permuted language modeling (PLM) for pre-training to capture dependencies between predicted tokens, thus overcoming the limitations of BERT. 
MPNet~\cite{DBLP:conf/nips/Song0QLL20} integrates MLM and PLM to inherit the advantages of both methods.

Furthermore, \cite{DBLP:journals/corr/abs-2202-08005} found that masking up to 40\% of input tokens can outperform the baseline of masking 15\% input tokens, measured by fine-tuning on downstream tasks. However, higher masking rates aggravates the mismatch problem.
In this paper, we introduce a simple and efficient DropMask method to alleviate the mismatch problem to improve the transfer learning ability of BERT without adding additional parameters and increasing training cost.

\section{Implementation Details}
\label{sec:implementation}
We utilize the HuggingFace toolkit\footnote{https://github.com/huggingface/transformers} to implement all models and DropMask method and conduct experiments.
For Chinese pre-training, we use Masked Language Modeling (MLM)~\cite{DBLP:conf/naacl/DevlinCLT19} and Sentence Structure Objective (SSO)~\cite{DBLP:conf/iclr/0225BYWXBPS20} on Chinese Wikipedia (3GB text data)\footnote{https://dumps.wikimedia.org/}.
% We use Adam \cite{} as the optimizer. 
The masking rate is 15\%.
We train for 100,000 steps with a learning rate of 1e-4, a batch size of 384, and a maximum length of 512. For fine-tuning on CLUE, we run 5 runs with different random seeds and compute the mean and standard deviation.
For each seed, we use a grid search on hyperparameters to obtain the best results for the seed.
For AFQMC, CSL, iFLYTEK, OCNLI, TNEWS, hyperparameters are searched by batch size in \{16, 32, 48\} and learning rate in \{1e-5, 3e-5, 5e-5\}, the number of epochs is 5; For WSC, hyperparameters are  searched by batch size in \{16, 32, 48\}, learning rate in \{1e-5, 2e-5, 3e-5, 4e-5, 5e-5\}, and the number of epochs in \{50, 100\}; for CMNLI, batch size is 48, learning rate is 1e-5, and the number of epochs is 5. 

For English pre-training, we only use MLM on English Wikipedia and BooksCorpus (16GB text data)\footnote{https://huggingface.co/datasets/bookcorpus}.
We train for 125,000 steps with a learning rate of 4e-4, a batch size of 2048, and a maximum length of 128. For fine-tuning on GLUE, we use grid search on hyperparameters to obtain the best results. Hyperparameters are searched in batch size 32, learning rate \{2e-5, 3e-5, 5e-5\}, number of epochs \{5, 10\}.
All models are comparable in the BERT-base size.

\end{document}